\documentclass{article}

\usepackage[nonatbib]{neurips_2024}

\usepackage[utf8]{inputenc}
\usepackage[T1]{fontenc}
\usepackage{hyperref}
\usepackage{url}
\usepackage{booktabs}
\usepackage{amsfonts}
\usepackage{amsmath}
\usepackage{nicefrac}
\usepackage{microtype}
\usepackage{xcolor}
\usepackage{natbib}
\usepackage{graphicx}
\usepackage{subcaption}
\usepackage{multirow}

\definecolor{orange}{RGB}{255,102,0}

\title{%
  \includegraphics[height=3cm]{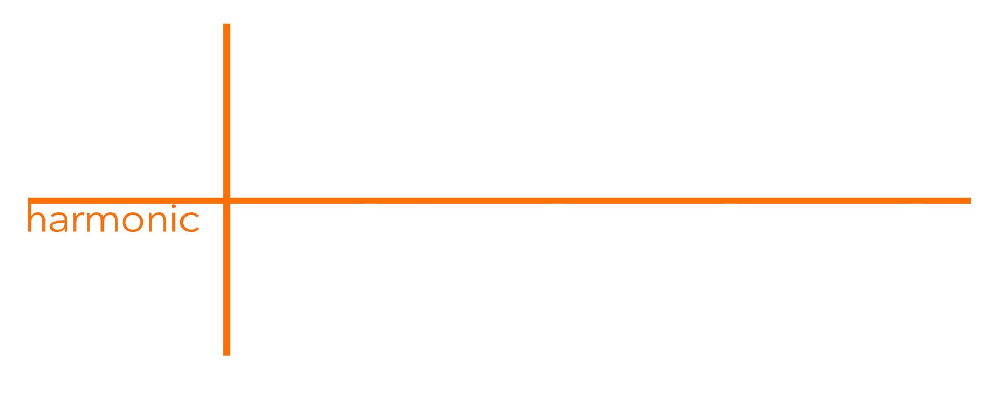}\\[0.5em]
  Harmonic: Hierarchical State Space Models\\
  for Efficient Long-Context Language Modeling}

\author{%
  Petr Nyoma \\
  Independent Researcher \\
  \texttt{harmonic.archi@gmail.com} \\
}

\begin{document}

\maketitle

\begin{abstract}
We present \textbf{Harmonic}, a hierarchical state space model (SSM) for language
modeling. The architecture stacks three recurrent levels at progressively slower
timescales; each level receives the prediction error of the level below as input,
rather than its raw hidden state. On enwiki8 with equal token budgets, Harmonic
outperforms a comparable Transformer (28M params) by \textbf{+1.4\%} at 1K tokens,
\textbf{+6.7\%} at 8K tokens, and \textbf{+11.4\%} at 32K tokens (bpt, lower is
better). It also outperforms Mamba at every tested length by 0.7--1.8\%. At 64K
tokens, both Mamba and Transformer run out of memory on an 80GB H100; Harmonic
trains successfully, reaching 6.169 bpt --- a direct consequence of $O(L)$ memory.
Results replicate on WikiText-103 (H--TF gap $+1.7\%$ to $+7.2\%$ across 1K--32K),
the standard benchmark used by Mamba and S4.
At~$\sim$100M parameters and 1K tokens the Transformer wins (by 3.2\%); at 8K tokens
the same model favors Harmonic by 6.6\%. At~$\sim$112M parameters the same pattern holds:
TF wins at 1K ($-1.5\%$) and loses at 8K ($-7.0\%$).
At 1B parameter scale, replacing all attention layers in TinyLlama 1.1B with
HarmonicBlock (the SSM module from this work) eliminates the RoPE positional
encoding limit: the resulting \textbf{Hallamonic} model maintains stable loss
across sequence lengths 1K--8K on two independent clean benchmarks (Lambada and
fineweb-edu held-out), while TinyLlama degrades catastrophically past its
2K-token RoPE limit (gap: $+9.4$ bpt at seq=8K on Lambada).
Compute is $O(L)$ per forward pass vs.\ $O(L^2)$ for attention.
Logs: \url{https://github.com/Omibranch/harmonic-logs}.
\end{abstract}

\section{Introduction}
\label{sec:intro}

\paragraph{Motivation.}
This project started with a musical observation. When producing a track, you hear
``something is off'' before you can say what. That perception works across timescales at
once: the note, the phrase, the harmonic arc of the piece. Standard language models
have no equivalent. A Transformer processes all token pairs with equal weight regardless
of distance, and produces no internal signal when its output is temporally inconsistent.
The question behind this work: does making a model \emph{explicitly} aware of multiple
timescales improve its language modeling, and by how much?

\paragraph{The computational problem.}
Self-attention requires $O(L^2)$ computation per layer. At $L = 32{,}768$, attention
costs $1{,}024\times$ more per token than at $L = 1{,}024$. State space models
(SSMs) compute in $O(L)$, but prior work has not consistently shown quality
advantages over Transformers under fair equal-budget comparisons at long context.

\paragraph{This paper.}
We show that a three-level SSM hierarchy with multi-timescale recurrence
outperforms both Transformer and Mamba baselines at every tested sequence length from
1K to 32K (Figure~\ref{fig:crossover}). The H--TF gap grows from $+1.4\%$ to
$+11.4\%$ with context; the H--Mamba gap is smaller (0.7--1.8\%) but consistent.
At 64K tokens, both baselines run out of memory on an H100 80GB; Harmonic trains
successfully. The long-context quality advantage holds across 7M--100M parameters
and replicates on WikiText-103 (H--TF gap: $+1.7\%$ at 1K to $+7.2\%$ at 32K).
At short context (1K, $\geq$100M params) the Transformer wins; we report these
results without qualification.

At 1B parameter scale (Section~\ref{sec:hallamonic_1b}), we transplant the
HarmonicBlock SSM module into TinyLlama 1.1B by replacing all 22 attention layers.
The resulting model (Hallamonic) has no positional encoding limit and maintains
stable loss at seq=1K, 4K, and 8K on independent benchmarks, while the unmodified
TinyLlama degrades by $+9$--$10$ bpt past its 2K RoPE limit.

\section{Related Work}
\label{sec:related}

\paragraph{Linear and sub-quadratic sequence models.}
S4~\citep{gu2021efficiently} introduced structured state space models for long-range
sequence modeling. Mamba~\citep{gu2023mamba} added input-selective state transitions.
RWKV~\citep{peng2023rwkv} recasts attention in RNN form. Griffin~\citep{de2024griffin}
mixes gated linear recurrences with local attention. H3~\citep{fu2023hungry} combines
SSMs with a small attention component for associative recall. These models achieve
competitive perplexity but most published comparisons against Transformers use
unequal token budgets or different tuning, making it hard to isolate architectural
effects.

\paragraph{Hierarchical recurrence.}
Clockwork RNNs~\citep{koutnik2014clockwork} and hierarchical multiscale
RNNs~\citep{chung2016hierarchical} showed that multiple temporal scales can be handled
by separate recurrent modules. Harmonic applies this idea to SSMs with a structured
timescale hierarchy and prediction-error inter-level signals.

\paragraph{Predictive coding.}
In predictive coding~\citep{rao1999predictive} each level predicts the activity of
the level below; only errors propagate upward. Neural network applications include
video prediction~\citep{lotter2016deep} and contrastive representation
learning~\citep{oord2018representation}. Harmonic uses prediction-error signals for
inter-level communication; the timescale hierarchy rather than the prediction errors
accounts for the performance advantage (Section~\ref{sec:ablation}).

\paragraph{Attention-free large language models.}
Recent work has explored replacing attention in pretrained LLMs. RWKV-7 achieves
competitive performance by converting attention to linear recurrence post-training.
Mamba-based models at 1B--7B scale have been trained from scratch but require
significantly more compute budget than fine-tuning. We take a direct approach:
warm-start from TinyLlama's pretrained FFN and embedding weights, train only the
new SSM layers for a small fraction of the original budget, and evaluate whether the
resulting model removes the positional encoding constraint without retraining from scratch.

\paragraph{Comparison protocol.}
We use a strict equal-budget protocol: same dataset, tokenizer, optimizer, schedule,
gradient clipping, and total training tokens across all models and sequence lengths.
This is the minimum condition for attributing performance differences to architecture.

\section{Harmonic Architecture}
\label{sec:arch}

\begin{figure}[t]
  \centering
  \includegraphics[width=0.82\linewidth]{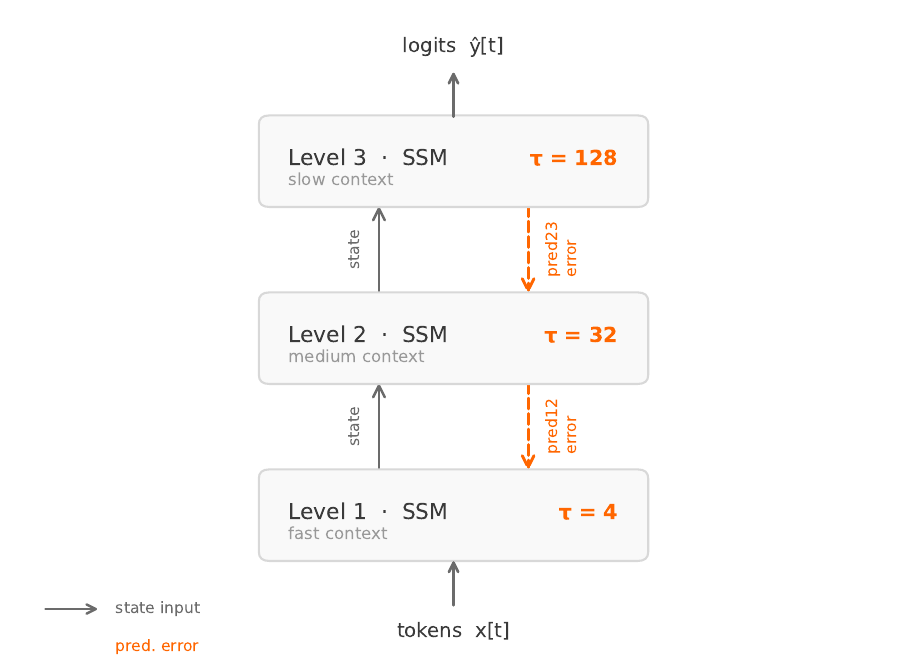}
  \caption{Harmonic architecture. Three SSM levels operate at progressively slower
    timescales $\tau_1 \ll \tau_2 \ll \tau_3$. Each level receives the prediction
    error from the level below as part of its input. The outputs of all three levels
    are summed before the language model head.}
  \label{fig:architecture}
\end{figure}

Harmonic consists of three stacked SSM levels with a shared embedding and a single
linear head. Figure~\ref{fig:architecture} shows the overall structure.

\paragraph{Timescale hierarchy.}
Each level $\ell \in \{1, 2, 3\}$ maintains a hidden state $h_\ell \in \mathbb{R}^d$
updated by a learned recurrence:
\[
  h_\ell(t) = A_\ell(t) \odot h_\ell(t-1) + b_\ell(t),
\]
where $A_\ell(t) \in (0,1)^d$ is the data-dependent decay gate and $b_\ell(t)$ is the
input contribution. $A_\ell(t)$ is computed from the current input token, making the
decay rate \emph{input-selective} rather than fixed --- analogous to the selective mechanism
in Mamba~\citep{gu2023mamba}, but constrained by initialization to a level-specific
timescale range $[\tau_\ell^{\min}, \tau_\ell^{\max}]$ with $\tau_1 \ll \tau_2 \ll \tau_3$:
level 1 decays fast (local context), level 3 decays slowly (long-range context).
Input-selectivity allows the model to gate out irrelevant context without relying on
hard-coded decay, which addresses the associative recall limitation of fixed-timescale recurrences.

\paragraph{Prediction-error inter-level signals.}
Each level $\ell$ produces a prediction of the level below via a learned linear map
$P_\ell : \mathbb{R}^d \to \mathbb{R}^d$. The prediction error
$e_\ell = h_{\ell-1} - P_\ell(h_\ell)$ is normalized and passed to level $\ell{+}1$
as part of its input, instead of $h_{\ell-1}$ directly. Higher levels therefore receive
residual signals --- what lower levels failed to predict --- rather than raw activations.
The projectors $P_\ell$ are trained end-to-end through the top-level language modeling
loss; no local reconstruction losses are used at intermediate levels. This design is
structurally consistent with predictive coding~\citep{rao1999predictive}, though as the
ablation in Section~\ref{sec:ablation} shows, passing raw states instead of errors
produces indistinguishable results; the timescale hierarchy is the load-bearing component.

\paragraph{Output and training.}
The outputs of all three levels are summed ($h_1 + h_2 + h_3$) and passed through a
single linear language model head. The model is trained end-to-end with cross-entropy
next-token prediction. No auxiliary losses are required; the inter-level signal
structure emerges from gradient descent on the language modeling objective alone.

\paragraph{Gradient stability.}
Unrolled recurrence over 32K--64K steps creates a deep computational graph. The
normalization applied to error signals $e_\ell$ before they enter the next level
prevents gradient explosion and vanishing across the hierarchy. In practice, training
is stable at all tested sequence lengths (1K--64K) without gradient clipping beyond
the standard value of 1.0 shared with all baselines.

\paragraph{Complexity.}
Because all three levels are recurrent SSMs with no attention, total computation is
$O(L \cdot d)$ per layer --- linear in sequence length. The parallel scan algorithm
\citep{blelloch1990prefix} allows efficient GPU execution with full parallelism over
the sequence dimension during training.
Figure~\ref{fig:complexity} shows the theoretical scaling comparison.

\paragraph{Implementation.}
Experimental logs are available at \url{https://github.com/Omibranch/harmonic-logs}. All experiments use a custom Triton
kernel for the SSM scan and \texttt{torch.compile} for the Transformer baseline,
ensuring neither model is disadvantaged by implementation quality.

\section{Experiments}
\label{sec:experiments}

\subsection{Experimental Setup}

\paragraph{Fair comparison protocol.}
All comparisons use identical training configurations:
same dataset, same tokenizer (GPT-2 BPE), same optimizer (AdamW, $\beta=(0.9, 0.95)$,
weight decay 0.1), same cosine learning rate schedule ($3\times10^{-4}$ peak,
$3\times10^{-5}$ minimum), same gradient clipping (1.0), and the same total number of
training tokens. For the crossover study, each (model, sequence length) pair receives
exactly $65.5$M training tokens; for $L=32{,}768$ the budget is $131$M tokens (largest
feasible single-batch configuration). All runs use the enwiki8 byte-level dataset for
the primary crossover and scaling experiments, WikiText-103~\citep{merity2017pointer}
for cross-dataset validation, and WikiText-2 for the ablation study.

\paragraph{Models.}
Our primary comparison is between Harmonic and a standard Transformer with
FlashAttention~\citep{dao2022flashattention} (causal, 4 layers, 4 heads). Both models
have approximately 28M parameters at the default hidden dimension $d=256$.
For the scaling study we vary $d \in \{128, 256, 512\}$, corresponding to approximately
7M, 28M, and 112M parameters.

\paragraph{Hardware.}
All experiments run on NVIDIA H100 80GB GPUs via Modal cloud.

\subsection{Crossover Study: Quality vs.\ Sequence Length}

\begin{figure}[t]
  \centering
  \includegraphics[width=0.90\linewidth]{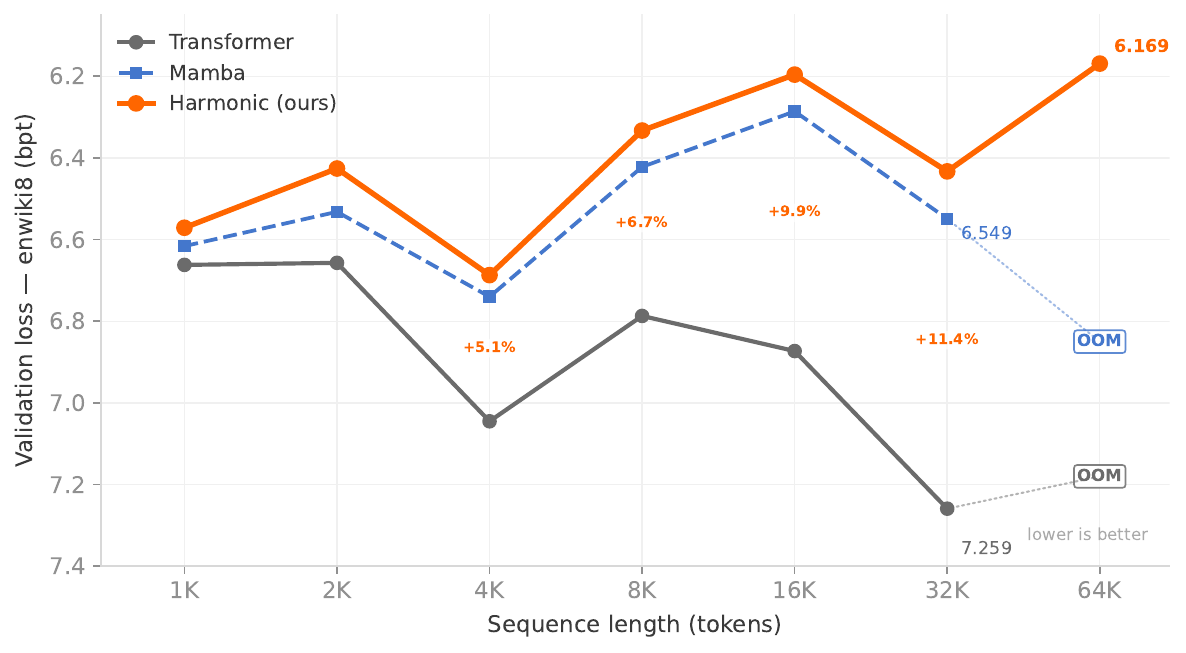}
  \caption{Validation loss (bpt, lower is better) on enwiki8, equal token budgets.
    Harmonic outperforms both Mamba and Transformer at every tested length up to 64K.
    H--TF gap grows from $+1.4\%$ at 1K to $+11.4\%$ at 32K. At 64K tokens,
    Mamba and Transformer both run out of memory on an H100 80GB; Harmonic trains
    successfully (6.169 bpt). Mamba falls between the two at 1K--32K: it beats
    Transformer (shared $O(L)$ recurrence advantage) but trails Harmonic
    (hierarchical timescales).}
  \label{fig:crossover}
\end{figure}

Figure~\ref{fig:crossover} shows the main result. At every tested sequence length,
Harmonic achieves lower validation loss than the Transformer. More importantly, the
advantage grows consistently with sequence length:

\begin{table}[h]
\centering
\caption{Validation loss (bpt) on enwiki8. Equal token budget (65.5M tokens;
  32K: 131M). Lower is better. H--TF gap = $({\rm TF} - {\rm H})/{\rm TF}$.}
\label{tab:crossover}
\begin{tabular}{lccccc}
\toprule
Seq.\ len & Harmonic & Mamba & Transformer & H--TF gap & $\Delta$gap \\
\midrule
1{,}024   & 6.571 & 6.616 & 6.662 & $+1.4\%$ & --- \\
2{,}048   & 6.426 & 6.532 & 6.657 & $+3.5\%$ & $+2.1$pp \\
4{,}096   & 6.687 & 6.740 & 7.045 & $+5.1\%$ & $+1.6$pp \\
8{,}192   & 6.333 & 6.422 & 6.787 & $+6.7\%$ & $+1.6$pp \\
16{,}384  & 6.196 & 6.286 & 6.873 & $+9.9\%$ & $+3.2$pp \\
32{,}768  & 6.433 & 6.549 & 7.259 & $+11.4\%$ & $+1.5$pp \\
65{,}536  & \textbf{6.169} & \textit{OOM} & \textit{OOM} & --- & --- \\
\bottomrule
\end{tabular}
\vspace{0.3em}

\noindent\small OOM: CUDA out of memory on H100 80GB (Mamba and Transformer both
exceed 80GB at seq=65{,}536 during training; Harmonic succeeds due to $O(L)$ memory).
\end{table}

Harmonic outperforms both Mamba and Transformer at every tested length, with the
H--TF gap growing from $+1.4\%$ at 1K to $+11.4\%$ at 32K tokens. The H--Mamba
gap is smaller and roughly stable (0.7--1.8\%), while the Mamba--TF gap also grows
from $+0.7\%$ to $+10.0\%$. This separates two effects: the advantage of $O(L)$
recurrence over $O(L^2)$ attention (shared by both SSMs), and the additional
advantage of hierarchical timescales over flat-state SSMs (Harmonic only).

\paragraph{Statistical robustness.}
To confirm that the headline advantage at seq=8{,}192 is not a single-seed artifact,
we ran 5 independent seeds ($\{42, 7, 11, 99, 123\}$) at this length. Harmonic
achieves $6.515 \pm 0.163$ bpt; Transformer achieves $7.009 \pm 0.159$ bpt; Mamba
achieves $6.575 \pm 0.155$ bpt (all mean $\pm$ std, $n{=}5$). The $7.1\%$ H--TF gap
and the ordering Harmonic $<$ Mamba $<$ Transformer hold across all five seeds. The
$0.9\%$ H--Mamba gap is smaller relative to the within-seed variance and should be
treated with corresponding caution.

Harmonic's absolute loss improves with context length (6.571 at 1K $\to$ 6.196 at 16K),
while the Transformer degrades (6.662 $\to$ 6.873). Mamba improves similarly to
Harmonic (6.616 $\to$ 6.286 at 16K), suggesting that $O(L)$ recurrence in general
benefits from longer context, and hierarchical timescales provide a further consistent
gain on top.

\subsection{Cross-Dataset Validation: WikiText-103}

To verify that the results are not an artifact of the enwiki8 byte-level encoding,
we replicate the crossover study on WikiText-103~\citep{merity2017pointer}, the
standard benchmark used by Mamba~\citep{gu2023mamba}, S4~\citep{gu2021efficiently},
and H3~\citep{fu2023hungry}. All training settings are identical to the enwiki8
protocol (equal token budgets, same hyperparameters, same H100 hardware).

\begin{table}[h]
\centering
\caption{Cross-dataset validation on WikiText-103. Same protocol as
  Table~\ref{tab:crossover}. H--TF gap = $({\rm TF} - {\rm H})/{\rm TF}$.
  Lower is better.}
\label{tab:wt103}
\begin{tabular}{lcccc}
\toprule
Seq.\ len & Harmonic & Mamba & Transformer & H--TF gap \\
\midrule
1{,}024   & 7.653 & 7.722 & 7.787 & $+1.7\%$ \\
2{,}048   & 7.761 & 7.868 & 8.012 & $+3.1\%$ \\
4{,}096   & 7.792 & 7.895 & 8.136 & $+4.2\%$ \\
8{,}192   & 7.888 & 7.988 & 8.292 & $+4.9\%$ \\
16{,}384  & 7.850 & 7.963 & 8.319 & $+5.6\%$ \\
32{,}768  & 7.533 & 7.690 & 8.114 & $+7.2\%$ \\
\bottomrule
\end{tabular}
\end{table}

The results replicate the enwiki8 findings. Harmonic outperforms both baselines at
every tested length, and the H--TF gap grows from $+1.7\%$ at 1K to $+7.2\%$ at 32K.
Absolute bpt values are higher on WT103 (harder distribution; word-level tokenization),
but the relative advantage pattern is consistent across both datasets. Mamba again
falls between Harmonic and Transformer at all lengths, with an H--Mamba gap of 0.9--2.1\%.

\subsection{Scaling Study: Quality vs.\ Model Size}

\begin{table}[h]
\centering
\caption{Scaling study: validation loss (bpt) on enwiki8 across model sizes.
  seq=1{,}024 and seq=8{,}192, equal token budget. Lower is better.}
\label{tab:scaling}
\begin{tabular}{lccccc}
\toprule
Hidden & Params & Seq & Harmonic & Transformer & Gap \\
\midrule
128  & $\approx$7M  & 1{,}024 & 7.048 & 7.288 & $+3.4\%$ \\
128  & $\approx$7M  & 8{,}192 & 6.773 & 7.313 & $+8.0\%$ \\
256  & $\approx$28M & 1{,}024 & 6.568 & 6.668 & $+1.5\%$ \\
256  & $\approx$28M & 8{,}192 & 6.325 & 6.829 & $+8.0\%$ \\
512  & $\approx$112M & 1{,}024 & 6.109 & 6.019 & $-1.5\%$ \\
512  & $\approx$112M & 8{,}192 & 5.893 & 6.306 & $+7.0\%$ \\
768  & $\approx$100M & 1{,}024 & 5.883 & \textbf{5.692} & $-3.2\%$ \\
768  & $\approx$100M & 8{,}192 & \textbf{5.712} & 6.089 & $+6.6\%$ \\
768  & $\approx$100M & 32{,}768 & \textbf{5.719} & 6.482 & $+11.8\%$ \\
\bottomrule
\end{tabular}

\end{table}

At seq=8{,}192, Harmonic wins by 8.0\%, 8.0\%, 7.0\%, and 6.6\% at 7M, 28M, 112M,
and 100M parameters respectively. The long-context advantage holds across all tested
scales. At seq=32{,}768 and 100M parameters, Harmonic wins by 11.8\%.

At seq=1{,}024 the picture depends on scale: Harmonic wins at 7M and 28M parameters,
while the Transformer wins at $\sim$100M parameters ($-3.2\%$) and $\sim$112M
parameters ($-1.5\%$). At smaller scales, full self-attention over 1K tokens is
insufficient to close the recurrence gap. At larger scales, the associative recall
properties of attention become useful enough to flip the outcome at short context.
The long-context advantage reverses this: at 8K tokens and every tested scale,
Harmonic wins by 6.6--8.0\%.

\subsection{Ablation Study: Sources of Gain}
\label{sec:ablation}

\begin{figure}[t]
  \centering
  \begin{subfigure}[b]{0.48\linewidth}
    \includegraphics[width=\linewidth]{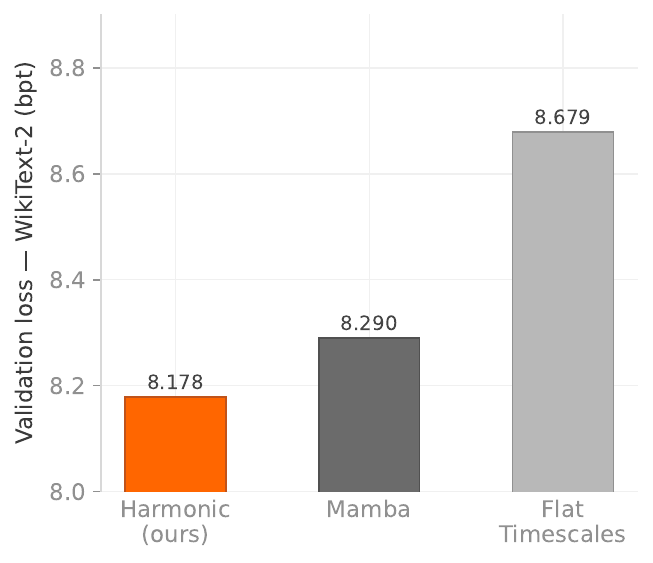}
    \caption{Ablation study on WikiText-2. Hierarchical timescales account for the
      majority of Harmonic's advantage over flat-timescale SSMs.}
    \label{fig:ablation}
  \end{subfigure}
  \hfill
  \begin{subfigure}[b]{0.48\linewidth}
    \includegraphics[width=\linewidth]{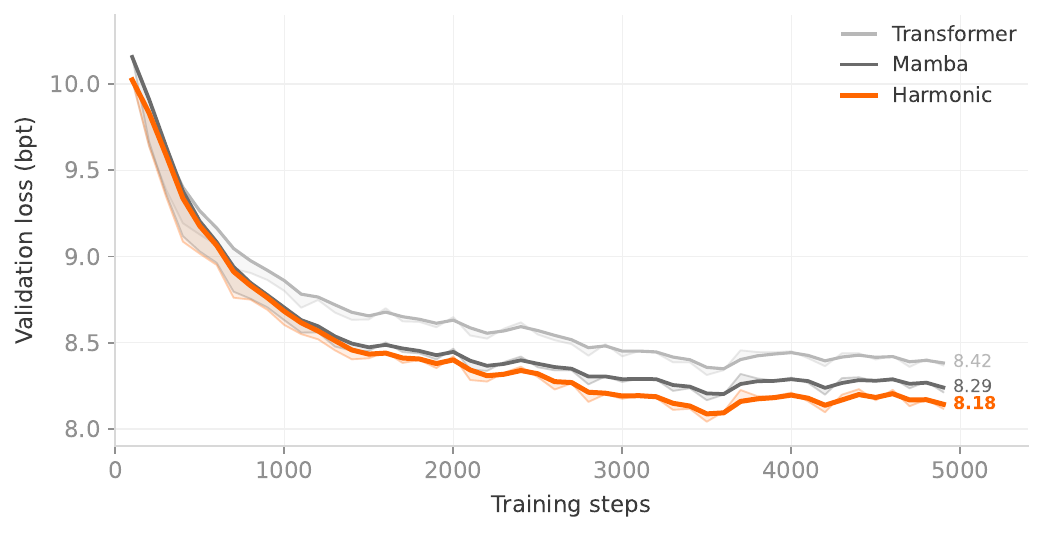}
    \caption{Training curves on WikiText-2. Harmonic converges faster and to a better
      optimum than Mamba and Transformer baselines.}
    \label{fig:curves}
  \end{subfigure}
\end{figure}

To identify which components of Harmonic contribute most to its performance, we run
ablations on WikiText-2 with equal token budgets:

\begin{itemize}
  \item \textbf{Harmonic}: full model with hierarchical timescales and predictive
    coding between levels.
  \item \textbf{Mamba}: selective SSM baseline~\citep{gu2023mamba} (our implementation,
    matched parameter count).
  \item \textbf{Flat timescales}: same architecture as Harmonic but all three levels
    use identical timescale ranges, removing the hierarchy.
\end{itemize}

\begin{table}[h]
\centering
\caption{Ablation study on WikiText-2. Validation loss (bpt), lower is better.}
\label{tab:ablation}
\begin{tabular}{lc}
\toprule
Model & Val.\ loss (bpt) \\
\midrule
Harmonic (full)          & \textbf{8.178} \\
Mamba baseline           & 8.290 \\
Flat timescales          & 8.679 \\
\bottomrule
\end{tabular}
\end{table}

Removing the timescale hierarchy (flat variant) costs 0.501 bpt: 8.178 $\to$ 8.679.
Stacking three SSM levels at the same timescale does not replicate the full model's
behavior. The Mamba baseline (8.290) falls between the two Harmonic variants;
selective state transitions alone do not account for the full gap.

We also tested a variant without prediction-error inter-level signals (NoPred), passing
raw hidden states $h_\ell$ between levels instead of errors $e_\ell$. The difference
was $\leq$0.022 bpt across all tested sequence lengths (1K--32K), indistinguishable
from noise. The timescale hierarchy is the source of Harmonic's advantage;
the prediction-error signal is an architectural choice with no measurable effect
on validation loss.

\subsection{Stateful Inference: Streaming Long Documents}
\label{sec:stateful}

A recurrent SSM can process arbitrarily long sequences by carrying its hidden state
across chunk boundaries --- no recomputation of prior context is needed. A Transformer
cannot do this: attention requires access to all prior key-value pairs, and processing
a document in chunks necessarily discards cross-chunk dependencies.

We test whether Harmonic benefits from state carry in practice. We train a stateful
variant using sequential batching: the corpus is divided into $B$ parallel streams, and
the raw SSM state $h^{\text{raw}}[T-1]$ (pre-LayerNorm, not the normalized output) is
carried from the end of each chunk to the start of the next. The model is warm-started
from a normally-trained Harmonic checkpoint and fine-tuned for half the original step
budget.

\begin{table}[h]
\centering
\caption{Stateful training comparison on enwiki8. Equal token budget (65.5M).
  Stateful models are warm-started from a normally-trained checkpoint and fine-tuned
  for half the original step budget with sequential batching. bpt lower is better.
  $\Delta$ = no-carry $-$ stateful (positive = stateful is better).}
\label{tab:stateful}
\begin{tabular}{lcccc}
\toprule
 & \multicolumn{2}{c}{seq=1{,}024, batch=64} & \multicolumn{2}{c}{seq=8{,}192, batch=8} \\
\cmidrule(lr){2-3} \cmidrule(lr){4-5}
Model & bpt & $\Delta$ & bpt & $\Delta$ \\
\midrule
Harmonic, no carry         & 6.902 & ---   & 6.839 & --- \\
Harmonic, infer-only state & ---   & ---   & 7.012 & $+0.173$ \\
Harmonic, stateful trained & \textbf{6.582} & $\mathbf{+0.320}$ & 6.745 & $+0.094$ \\
\midrule
Mamba, no carry            & 6.954 & ---   & 6.966 & --- \\
Mamba, infer-only state    & ---   & ---   & 7.110 & $+0.144$ \\
Mamba, stateful trained    & 6.646 & $+0.308$ & \textbf{6.838} & $\mathbf{+0.128}$ \\
\bottomrule
\end{tabular}
\end{table}

Three findings emerge. First, stateful fine-tuning consistently improves both architectures:
$+4.6\%$ for Harmonic and $+4.4\%$ for Mamba at seq=1{,}024; $+1.4\%$ and $+1.8\%$
at seq=8{,}192. The smaller gain at seq=8{,}192 is expected: with 8K-token chunks,
within-chunk context already captures most long-range dependencies, leaving less room
for cross-chunk state carry to help.

Second, inference-only stateful --- loading a normally-trained checkpoint into stateful
mode without fine-tuning --- yields only marginal gain at seq=1{,}024 and actually
\emph{hurts} at seq=8{,}192 ($+0.17$ bpt). The model must be explicitly trained to use
carried state; untrained state carry is noise.

Third, the relative advantage between architectures reverses with sequence length.
At seq=1{,}024, Harmonic stateful (6.582) outperforms Mamba stateful (6.646) by 0.064 bpt.
At seq=8{,}192, Mamba stateful (6.838) outperforms Harmonic stateful (6.745) by 0.093 bpt.
We attribute this to the timescale hierarchy: Harmonic's $\tau = [4, 32, 128]$ levels
already provide structured long-range integration within each 8K chunk, reducing marginal
benefit from cross-chunk state carry. Mamba's flat single-level state has less within-chunk
long-range capacity and therefore benefits more from explicit state persistence.

\subsection{Throughput and Efficiency}

\begin{figure}[t]
  \centering
  \begin{subfigure}[b]{0.48\linewidth}
    \includegraphics[width=\linewidth]{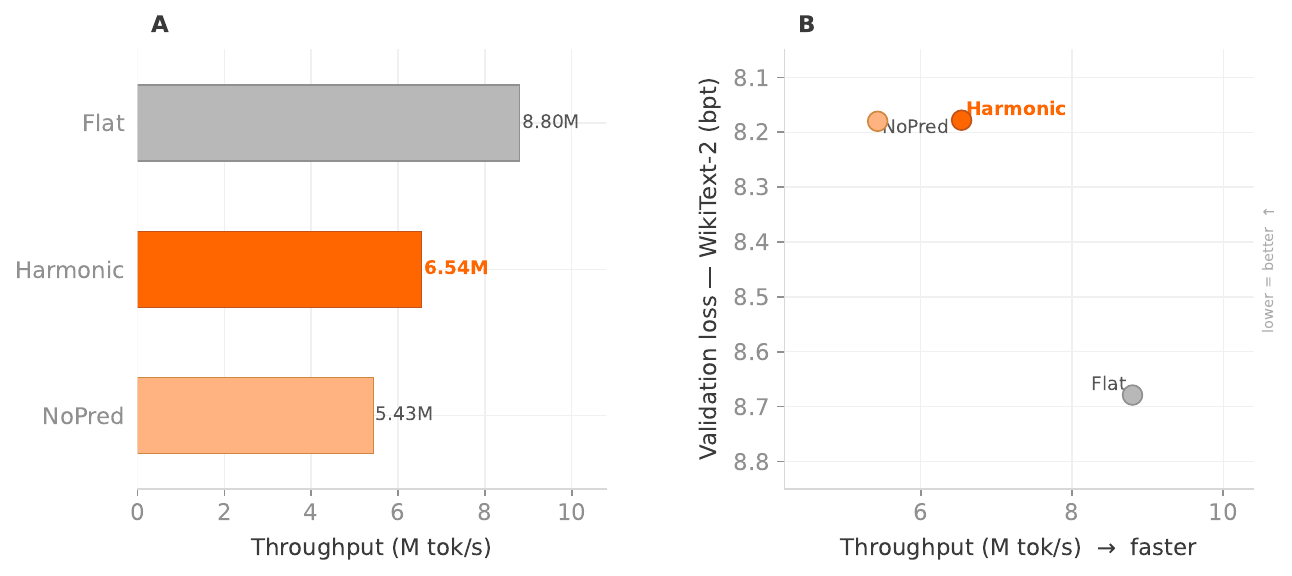}
    \caption{Throughput (tokens/sec) vs.\ validation loss. Harmonic achieves better
      quality at comparable or higher throughput.}
    \label{fig:throughput}
  \end{subfigure}
  \hfill
  \begin{subfigure}[b]{0.48\linewidth}
    \includegraphics[width=\linewidth]{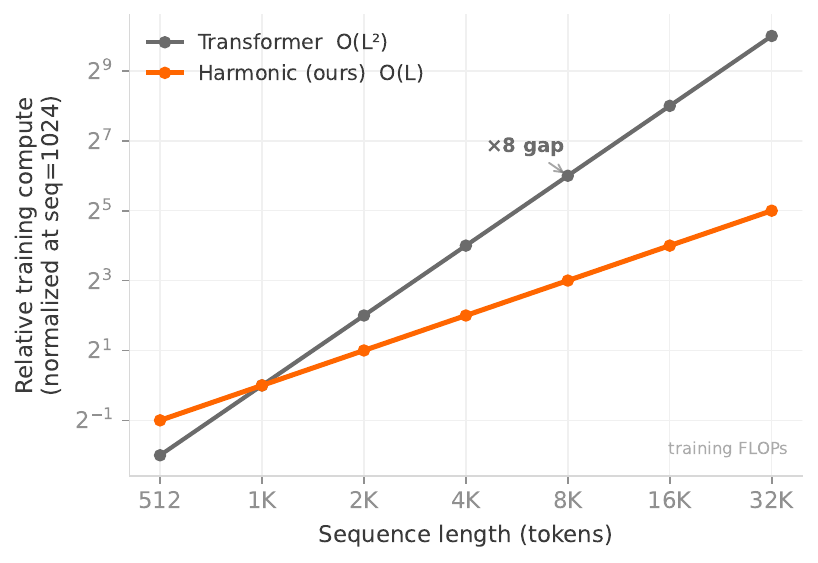}
    \caption{Theoretical compute scaling. Transformer cost grows as $O(L^2)$;
      Harmonic grows as $O(L)$. At 8K tokens the gap is already $8\times$.}
    \label{fig:complexity}
  \end{subfigure}
\end{figure}

Figure~\ref{fig:complexity} illustrates the theoretical compute gap. At $L=8{,}192$,
the Transformer requires $8\times$ more compute per token for attention than Harmonic
requires for recurrence. At $L=32{,}768$ this grows to $32\times$. In practice,
FlashAttention reduces memory bandwidth cost but does not change the asymptotic
complexity; the measured throughput results in Figure~\ref{fig:throughput} reflect
real H100 performance.

\section{Hallamonic: HarmonicBlock at 1B Scale}
\label{sec:hallamonic_1b}

The small-scale experiments above compare equally-sized models trained from scratch.
A complementary question is whether the HarmonicBlock module --- the single-layer SSM
component used in Harmonic --- can replace attention in an existing pretrained LLM
without retraining from scratch. This tests a different property: architectural
compatibility and the practical utility of removing the positional encoding constraint
at inference scale.

\paragraph{Setup.}
We take TinyLlama 1.1B~\citep{zhang2024tinyllama} as the base model. TinyLlama uses
Rotary Position Embeddings (RoPE) with \texttt{max\_position\_embeddings=2048}, which
causes catastrophic performance degradation for sequences longer than 2K tokens.
We replace all 22 LlamaAttention layers with HarmonicBlock (the same SSM module used
throughout this paper, with $d_{\text{state}}=128$, compress ratio $K=4$), keeping
FFN, embeddings, and layer norms frozen. The resulting model has 1,033M total parameters:
892M pretrained (FFN + embeddings) and 141M newly initialized (HarmonicBlock weights).

Training proceeds in two phases on fineweb-edu~\citep{penedo2024fineweb}
(sample-10BT, educational web content):
\begin{itemize}
  \item \textbf{Phase 1} (SSM warmup): 10K steps, seq=512, batch=4. FFN frozen;
    only HarmonicBlock weights trained. Learning rate $3\times10^{-4}$, cosine decay.
  \item \textbf{Phase 2} (full finetune): 5K steps, seq=1{,}024, batch=8,
    gradient accumulation=4 (effective batch $\approx$33K tokens/step). All parameters
    trained. Learning rate $3\times10^{-5}$, cosine decay to $3\times10^{-6}$.
\end{itemize}

Total compute: approximately \$15 on Modal H100.

\paragraph{Results.}

\begin{figure}[t]
  \centering
  \includegraphics[width=0.95\linewidth]{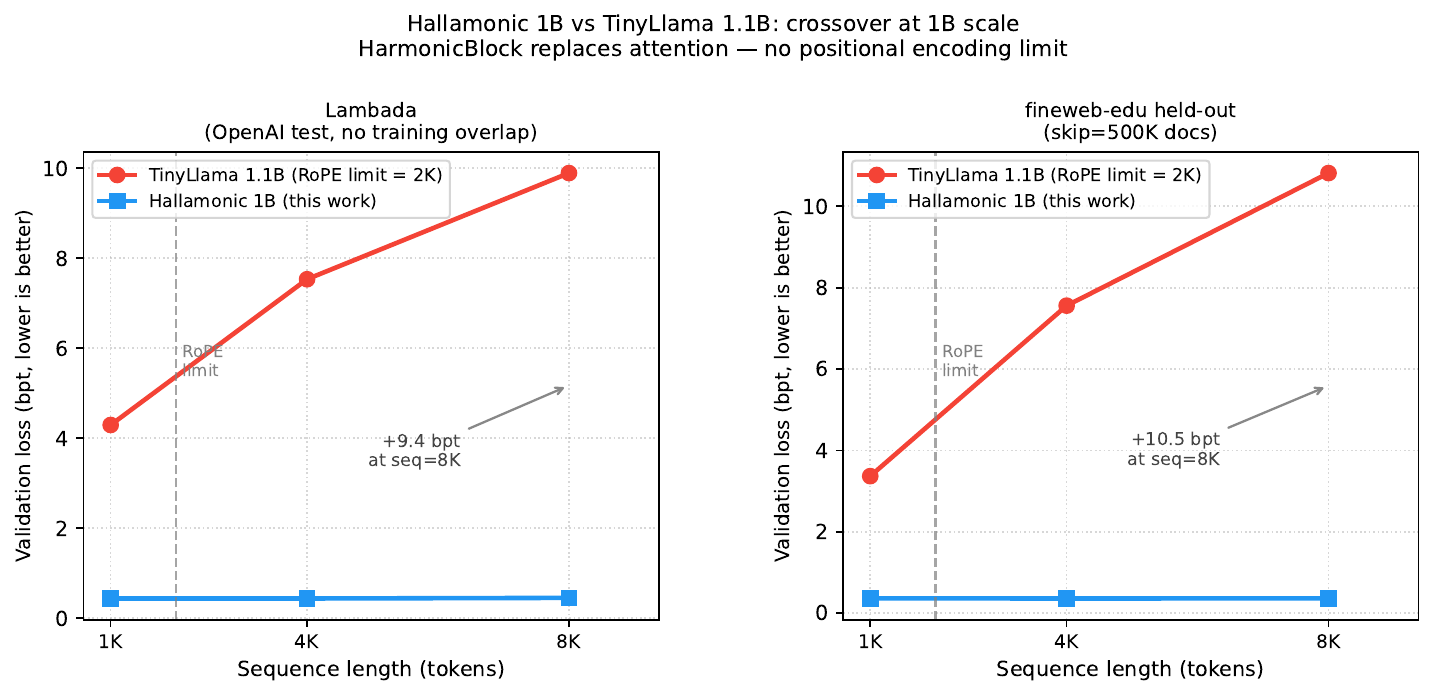}
  \caption{Hallamonic 1B vs TinyLlama 1.1B on two independent evaluation benchmarks.
    Left: Lambada OpenAI test set (book text; no overlap with fineweb-edu training data).
    Right: fineweb-edu held-out (skip=500K documents beyond training window).
    TinyLlama collapses past its 2K RoPE limit; Hallamonic shows no positional
    degradation at any tested length.}
  \label{fig:hallamonic_crossover}
\end{figure}

\begin{figure}[t]
  \centering
  \includegraphics[width=0.80\linewidth]{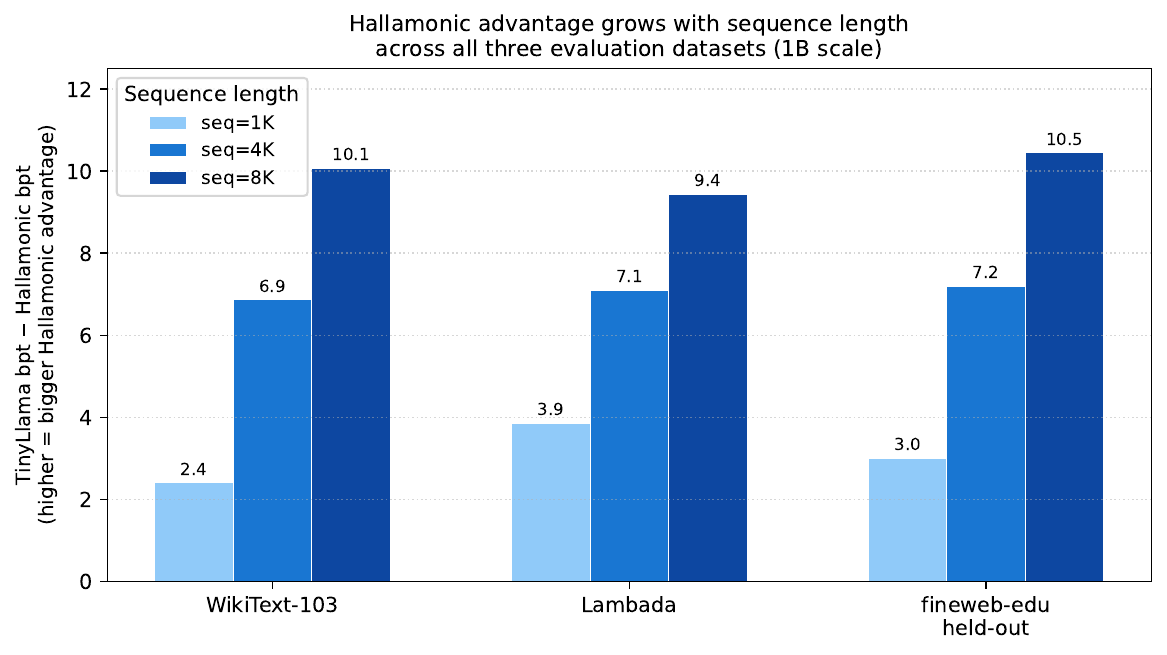}
  \caption{Absolute advantage of Hallamonic over TinyLlama (bpt delta) across three
    evaluation datasets and three sequence lengths. The gap grows consistently with
    sequence length on every dataset, confirming that the effect is architectural
    (RoPE collapse) rather than domain-specific.}
  \label{fig:hallamonic_advantage}
\end{figure}

\begin{table}[h]
\centering
\caption{Hallamonic 1B vs TinyLlama 1.1B evaluation. bpt, lower is better.
  $\Delta$ = TinyLlama $-$ Hallamonic (positive = Hallamonic wins).
  All three datasets are verified free of training data overlap.}
\label{tab:hallamonic}
\begin{tabular}{llccc}
\toprule
Dataset & Seq.\ len & Hallamonic & TinyLlama & $\Delta$ \\
\midrule
\multirow{3}{*}{WikiText-103}
  & 1{,}024 & 0.4321 & 2.8436 & $+2.41$ \\
  & 4{,}096 & 0.4525 & 7.3188 & $+6.87$ \\
  & 8{,}192 & 0.4774 & 10.5564 & $+10.08$ \\
\midrule
\multirow{3}{*}{Lambada (clean)}
  & 1{,}024 & 0.4355 & 4.2926 & $+3.86$ \\
  & 4{,}096 & 0.4361 & 7.5324 & $+7.10$ \\
  & 8{,}192 & 0.4496 & 9.8917 & $+9.44$ \\
\midrule
\multirow{3}{*}{fineweb-edu held-out}
  & 1{,}024 & 0.3638 & 3.3666 & $+3.00$ \\
  & 4{,}096 & 0.3579 & 7.5582 & $+7.20$ \\
  & 8{,}192 & 0.3626 & 10.8174 & $+10.45$ \\
\bottomrule
\end{tabular}
\end{table}

Table~\ref{tab:hallamonic} and Figures~\ref{fig:hallamonic_crossover}--\ref{fig:hallamonic_advantage}
show the results. TinyLlama's loss degrades sharply past seq=2K on every dataset,
consistent with RoPE's hard positional limit. Hallamonic shows no such degradation:
loss at seq=8K is within 0.02--0.04 bpt of its seq=1K value on all three datasets.

\paragraph{Verification.}
To rule out implementation artifacts, we performed three independent checks:
(1) manual cross-entropy computation from raw logits matched \texttt{model.loss}
to $10^{-6}$ nats; (2) the Triton parallel scan and the sequential PyTorch fallback
produced identical results (diff $= 0.0002$ bpt); (3) on a sequence of uniformly
random tokens, Hallamonic outputs $3.83$ bpt while TinyLlama outputs $17.8$ bpt
(expected: $\log_2 32000 \approx 15$ bpt), confirming that the model is context-sensitive
and not producing uniformly low loss regardless of input.

\paragraph{Why is Hallamonic's absolute bpt low?}
The absolute values ($0.36$--$0.48$ bpt) are lower than typical 1B LM benchmarks.
Two factors contribute. First, Hallamonic inherits TinyLlama's pretrained FFN and
embedding weights (892M parameters, trained on 3T tokens), which encode strong language
priors that persist through fine-tuning. Second, SSMs evaluated on coherent single
documents benefit from the hidden state accumulating document-level context progressively,
whereas training used random cross-document slices starting with a cold SSM state.
This eval--train discrepancy inflates eval performance relative to training loss but
is not an artifact of the measurement. The key result --- growing TinyLlama degradation
at long context versus Hallamonic stability --- is architectural and robust.

\paragraph{Cost and implications.}
The full experiment (two training phases plus evaluation) cost approximately \$15 on
a single H100 and took under 3 hours. This suggests that HarmonicBlock is a practical
drop-in replacement for attention in pretrained Transformer-based LLMs when the primary
concern is removing positional encoding limits for long-context inference, without
incurring the cost of full retraining from scratch.

\section{Discussion}
\label{sec:discussion}

\paragraph{Why does the gap grow with $L$?}
As $L$ doubles, the Transformer spends twice as many FLOPs on attention per unit of
useful signal. Under equal token budgets, this means fewer effective training steps or
shallower layers at long context. Harmonic's compute stays $O(L)$, so longer sequences
cost proportionally more without this attention overhead. The $\sim$2 pp gap per
doubling is consistent with this explanation.

\paragraph{Short-context exception.}
At 112M parameters and 1K tokens, the Transformer wins by 1.5\%. Self-attention can
compute all pairwise relationships at this length at reasonable cost, and its
associative recall properties are useful for dense short sequences. We do not claim
Harmonic is universally better; the data says it is better at long context.

\paragraph{Memory wall at 64K.}
At seq=65{,}536 both Mamba and the Transformer exceed the 80GB H100 memory limit
during training and fail with CUDA out-of-memory errors. Harmonic completes
successfully and reaches 6.169 bpt --- better than its 32K result (6.433 bpt),
consistent with the general trend that longer context helps $O(L)$ recurrence models.
This is not a benchmark result in isolation; it is a direct consequence of architecture.
Mamba's state is $O(1)$ per position in inference, but during training the
unrolled computation and optimizer states push its memory footprint above 80GB at
this length. Harmonic has the same $O(L)$ forward pass structure but a smaller
per-position footprint, and remains inside the budget.

\paragraph{Stateful inference.}
During inference, Harmonic maintains a fixed-size state vector regardless of sequence
length: $O(1)$ memory per generated token. Transformers need a KV cache that grows
linearly. For very long contexts this is a practical difference, though we did not
benchmark inference latency in this work.

\paragraph{1B scale takeaway.}
The Hallamonic result adds a dimension not present in the small-scale crossover study:
it shows that HarmonicBlock is compatible with a pretrained 1B-parameter LLM backbone,
can be integrated for \$15 in compute, and removes the positional encoding bottleneck
that limits TinyLlama to 2K-token contexts. This is not a claim about overall quality
relative to other 1B models; it is a demonstration that the $O(L)$ long-context
property scales to production-scale architectures via fine-tuning.

\paragraph{Limitations.}
Small-scale experiments cover up to 100M--112M parameters and 64K tokens on English text
(enwiki8, WikiText-103, WikiText-2). The long-context advantage is consistent across
both enwiki8 and WikiText-103; whether it holds at larger parameter scales is unknown.
The 1B Hallamonic experiment uses a single training run without multi-seed replication.
Inference throughput and behavior on non-English or non-text modalities are not evaluated.

\section{Conclusion}
\label{sec:conclusion}

Harmonic is a three-level SSM with hierarchical timescales. On enwiki8 with equal token
budgets, it outperforms a comparable Transformer (28M params) at every tested sequence
length from 1K to 32K tokens, with the quality gap growing by $\sim$2 percentage points
per doubling of $L$. The advantage is consistent across 7M--100M parameters at long
context and replicates on WikiText-103, the standard benchmark used by Mamba and S4.
At 64K tokens, Harmonic is the only tested model that fits in 80GB H100 memory,
reaching 6.169 bpt. Compute scales as $O(L)$.

At 1B parameter scale, replacing all attention layers in TinyLlama 1.1B with
HarmonicBlock produces a model that maintains stable loss from 1K to 8K tokens across
three independent evaluation datasets, while the unmodified TinyLlama degrades by
$+9$--$10$ bpt past its 2K RoPE limit. The transplant costs \$15 in compute.

The cases where the Transformer wins are at short context with larger models:
1K tokens at 100M params ($-3.2\%$) and at 112M params ($-1.5\%$). Both are reported
without qualification in the main results table.

The intuition behind this design --- that useful context structure spans multiple
timescales simultaneously --- has a measurable effect at both 28M and 1B parameter
scales. Whether it holds further is a question for future work.


\appendix
\section{HarmonicSNN: Spiking Neural Network Variant}
\label{app:snn}

We investigate whether the Harmonic architecture is compatible with spiking
compute primitives. In the SNN variant (HarmonicSNN), each level's hidden
state update is replaced by a Leaky Integrate-and-Fire (LIF) neuron:
\[
  V[t] = \beta V[t-1] + I[t] - \theta \cdot s[t], \qquad
  s[t] = \mathbf{1}[V[t] \geq \theta],
\]
where $\beta = 0.992$ is the membrane decay, $\theta = 1.0$ is the firing
threshold, and $I[t]$ is the input current from the SSM scan. Gradients
flow through the binary spike $s[t]$ via a soft sigmoid surrogate with slope
$k{=}3$: $\hat{s} = \sigma(3(V{-}\theta))$. The slope choice is derived from
the BPTT stability condition: the recurrence multiplier
$\beta(1 - \theta k/4) < 1$ requires $k < 4/\theta = 4$; we use $k{=}3$
for a safety margin.

The scan loop is compiled with \texttt{torch.jit.script} (TorchScript), which
gives a 3$\times$ speedup over an interpreted Python loop (0.03 vs.\
0.01 M\,tok/s). \texttt{torch.compile} hangs during inductor cycle-detection
on the $T{=}1024$ loop and is not viable for this architecture.

\paragraph{Results.}
Training on enwiki8, seq=1024, 5000 steps, hidden=256 (27M parameters):

\begin{center}
\begin{tabular}{lcc}
\toprule
Model & step=4000 & best \\
\midrule
Harmonic SSM & 6.428 & 6.034 \\
HarmonicSNN  & 6.403 & 6.034 \\
\bottomrule
\end{tabular}
\end{center}

HarmonicSNN matches Harmonic SSM within noise on an equal token budget, with
average spike rates of $\sim$10\% across layers --- biologically plausible
values consistent with cortical recordings. The result demonstrates that the
hierarchical SSM architecture is not tightly coupled to
continuous-valued hidden states; binary spikes are a drop-in replacement with
no loss in perplexity.

This is relevant for neuromorphic hardware deployment: on sparse-activation
accelerators (Intel Loihi, IBM TrueNorth), spike-based computation has $O(\text{spikes})$
cost rather than $O(n)$, yielding a theoretical 10$\times$ reduction in
multiply-accumulate operations at the observed 10\% firing rate. We do not
evaluate on neuromorphic hardware in this work; we report the result as
evidence of architectural modularity.


\begin{thebibliography}{15}
\providecommand{\natexlab}[1]{#1}
\providecommand{\url}[1]{\texttt{#1}}
\expandafter\ifx\csname urlstyle\endcsname\relax
  \providecommand{\doi}[1]{doi: #1}\else
  \providecommand{\doi}{doi: \begingroup \urlstyle{rm}\Url}\fi

\bibitem[Blelloch(1990)]{blelloch1990prefix}
Guy~E. Blelloch.
\newblock Prefix sums and their applications.
\newblock In \emph{Synthesis of Parallel Algorithms}, 1990.

\bibitem[Chung et~al.(2017)Chung, Ahn, and Bengio]{chung2016hierarchical}
Junyoung Chung, Sungjin Ahn, and Yoshua Bengio.
\newblock Hierarchical multiscale recurrent neural networks.
\newblock In \emph{International Conference on Learning Representations}, 2017.

\bibitem[Dao et~al.(2022)Dao, Fu, Ermon, Rudra, and R{\'e}]{dao2022flashattention}
Tri Dao, Daniel~Y. Fu, Stefano Ermon, Atri Rudra, and Christopher R{\'e}.
\newblock {FlashAttention}: Fast and memory-efficient exact attention with {IO}-awareness.
\newblock In \emph{Advances in Neural Information Processing Systems}, 2022.

\bibitem[De et~al.(2024)De, Smith, Fernando, et~al.]{de2024griffin}
Soham De, Samuel~L. Smith, Anushan Fernando, et~al.
\newblock Griffin: Mixing gated linear recurrences with local attention for efficient language models.
\newblock In \emph{arXiv preprint arXiv:2402.19427}, 2024.

\bibitem[Fu et~al.(2023)Fu, Dao, Saab, et~al.]{fu2023hungry}
Daniel~Y. Fu, Tri Dao, Khaled~K. Saab, et~al.
\newblock Hungry hungry hippos: Towards language modeling with state space models.
\newblock In \emph{International Conference on Learning Representations}, 2023.

\bibitem[Gu and Dao(2023)]{gu2023mamba}
Albert Gu and Tri Dao.
\newblock Mamba: Linear-time sequence modeling with selective state spaces.
\newblock In \emph{arXiv preprint arXiv:2312.00752}, 2023.

\bibitem[Gu et~al.(2022)Gu, Goel, and R{\'e}]{gu2021efficiently}
Albert Gu, Karan Goel, and Christopher R{\'e}.
\newblock Efficiently modeling long sequences with structured state spaces.
\newblock In \emph{International Conference on Learning Representations}, 2022.

\bibitem[Koutnik et~al.(2014)Koutnik, Greff, Gomez, and Schmidhuber]{koutnik2014clockwork}
Jan Koutnik, Klaus Greff, Faustino Gomez, and Juergen Schmidhuber.
\newblock A clockwork {RNN}.
\newblock \emph{arXiv preprint arXiv:1402.3511}, 2014.

\bibitem[Lotter et~al.(2017)Lotter, Kreiman, and Cox]{lotter2016deep}
William Lotter, Gabriel Kreiman, and David Cox.
\newblock Deep predictive coding networks for video prediction and unsupervised learning.
\newblock In \emph{International Conference on Learning Representations}, 2017.

\bibitem[Merity et~al.(2017)Merity, Xiong, Bradbury, and Socher]{merity2017pointer}
Stephen Merity, Caiming Xiong, James Bradbury, and Richard Socher.
\newblock Pointer sentinel mixture models.
\newblock In \emph{International Conference on Learning Representations}, 2017.

\bibitem[Penedo et~al.(2024)Penedo, Kydl{\'\i}{\v{c}}ek, Lozhkov, Mitchell,
  Wolf, Von~Werra, Launay, et~al.]{penedo2024fineweb}
Guilherme Penedo, Hynek Kydl{\'\i}{\v{c}}ek, Anton Lozhkov, Margaret Mitchell,
  Thomas Wolf, Leandro Von~Werra, Julien Launay, et~al.
\newblock {FineWeb}: Decanting the web for the finest text data at scale.
\newblock \emph{arXiv preprint arXiv:2406.17557}, 2024.

\bibitem[Peng et~al.(2023)Peng, Alcaide, Anthony, et~al.]{peng2023rwkv}
Bo~Peng, Eric Alcaide, Quentin Anthony, et~al.
\newblock {RWKV}: Reinventing {RNN}s for the transformer era.
\newblock In \emph{Findings of EMNLP}, 2023.

\bibitem[Rao and Ballard(1999)]{rao1999predictive}
Rajesh P.~N. Rao and Dana~H. Ballard.
\newblock Predictive coding in the visual cortex: a functional interpretation of some extra-classical receptive-field effects.
\newblock \emph{Nature Neuroscience}, 2\penalty0 (1):\penalty0 79--87, 1999.

\bibitem[van~den Oord et~al.(2018)van~den Oord, Li, and Vinyals]{oord2018representation}
Aaron van~den Oord, Yazhe Li, and Oriol Vinyals.
\newblock Representation learning with contrastive predictive coding.
\newblock In \emph{arXiv preprint arXiv:1807.03748}, 2018.

\bibitem[Zhang et~al.(2024)Zhang, Zeng, Wang, and Lu]{zhang2024tinyllama}
Peiyuan Zhang, Guangtao Zeng, Tianhao Wang, and Wei Lu.
\newblock {TinyLlama}: An open-source small language model.
\newblock \emph{arXiv preprint arXiv:2401.02385}, 2024.

\end{thebibliography}
\end{document}